\documentclass[runningheads,a4paper]{llncs}
\pdfoutput=1
\usepackage{amssymb}
\usepackage{amsmath}
\usepackage{graphicx}
\usepackage{url}
\usepackage{adjustbox}
\usepackage{array}
\usepackage[labelformat=empty]{subfig}
\usepackage[hyperindex,pdftex]{hyperref}
\usepackage[colorinlistoftodos]{todonotes}
\usepackage{booktabs} 
\usepackage{rotating}
\usepackage{multirow}
\usepackage{transparent}
\usepackage{tablefootnote}

\usepackage[subtle,margins=normal,leading=normal]{savetrees} 
\usepackage{bm} 
\usepackage{colortbl}
\usepackage{tabulary}

\begin{document}
\mainmatter              

\title{Weakly supervised segmentation\\ from extreme points}
\titlerunning{Weakly supervised segmentation from extreme points}  %

\author{
Holger Roth, Ling Zhang, Dong Yang, Fausto Milletari, \\
Ziyue Xu, Xiaosong Wang, Daguang Xu
}

\institute{
NVIDIA\thanks{Contact: \texttt{\{hroth,lingz,dongy,fmilletari,ziyuex,xiaosongw,daguangx\}@nvidia.com}}
}
\authorrunning{Holger Roth et al.}

\maketitle              

\begin{abstract}
Annotation of medical images has been a major bottleneck for the development of accurate and robust machine learning models. Annotation is costly and time-consuming and typically requires expert knowledge, especially in the medical domain. Here, we propose to use minimal user interaction in the form of extreme point clicks in order to train a segmentation model that can, in turn, be used to speed up the annotation of medical images. We use extreme points in each dimension of a 3D medical image to constrain an initial segmentation based on the random walker algorithm. This segmentation is then used as a weak supervisory signal to train a fully convolutional network that can segment the organ of interest based on the provided user clicks. We show that the network's predictions can be refined through several iterations of training and prediction using the same weakly annotated data. Ultimately, our method has the potential to speed up the generation process of new training datasets for the development of new machine learning and deep learning-based models for, but not exclusively, medical image analysis.
\end{abstract}

\section{Introduction}
\label{sec:intro}
The growing number of medical images taken in routine clinical practice increases the demand for machine learning (ML) methods to improve image analysis workflows. However, a major bottleneck for the development of novel ML-based models to integrate and increase the productivity of clinical workflows is the annotation of datasets that are useful to train such models.
At the same time, volumetric analysis has shown several advantages over 2D measurements for clinical applications \cite{devaraj2017nodulevolume}, which further increases the amount of data (a typical CT scan contains hundreds of slices) needing to be annotated in order to train accurate 3D models. 
However, the majority of annotation tools available today for medical imaging are constrained to annotation in multiplanar reformatted views. The annotator needs to either brush paint or draw boundaries around organs of interest, often on a slice-by-slice basis. Classical techniques like 3D region growing or interpolation tools can speed up the annotation process by starting from seed points or allowing the user to skip certain slices. However, their usability is often limited to certain types of structures and might not work well in general. 

Here, we propose to use minimal user interaction in form of extreme point clicks, together with iterative training and refinement. Starting from user-defined extreme points in each dimension of a 3D medical image, an initial segmentation is produced based on the random walker algorithm. This segmentation is then used as a weak supervisory signal to train a fully convolutional network that can segment the organ of interest based on the provided user clicks. We show that the network’s predictions can be iteratively refined by using several iterations of training and prediction using the same weakly annotated data.

\textbf{\textit{Related work:}}
Fully convolutional networks (FCNs) \cite{long2015fully} have established themselves as the state-of-the-art methods for medical image segmentation in recent years \cite{ronneberger2015u,milletari2016v,cciccek20163d}. However, a major drawback is that they are very data hungry, limiting their application in healthcare where data annotation is very expensive. In order to reduce the cost of labeling, semi-automated/interactive and weakly supervised methods have been proposed in the literature.

Building on recent advances in deep learning (DL), several methods have been proposed to integrate it with interactive segmentation schemes. 
DL has been used in \cite{wang2018deepigeos} for the DeepIGeoS algorithms, which leverages geodesic distance transforms and user scribbles to allow interactive segmentation. Such a method does not exhibit robust performance when seeking segmentation for unseen object classes. An alternative method \cite{wang2018interactive} uses image-specific fine-tuning and leveraging both bounding boxes and scribble-based interaction. In \cite{sakinis2019interactive}, the authors utilize point clicks that are modeled as Gaussian kernels in a separate input channel to a segmentation FCN in order to model user interactions via seed-point placing. Finally \cite{can2018learning} proposes to use user-provided scribbles with random walks \cite{grady2006random} and FCN predictions to achieve semi-automated segmentation of cardiac CT images. This method differs from our proposed method in that we only expect the user to provide extreme points rather than scribbles as initial input to the random walker algorithm and uses a different approach when iteratively refining the segmentations.

One of the first approaches using bounding box based weakly supervised training of deep neural networks in medical imaging was by \cite{rajchl2017deepcut}. They used a patch-based classification CNN to segment brain and lung regions using an initial \textit{GrabCut} segmentation. After several rounds of predictions using CNN plus Dense CRF post-processing, the network's segmentation performance could be improved.
Weakly-supervised or self-learning in medical image analysis can also make use of measurements readily available in the hospital picture archiving and communication system (PACS) such as measurements acquired during evaluation of the RECIST criteria \cite{cai2018accurate}. However, these measurements are typically constraint to 2D and might miss adequate constraints for more complex three-dimensional shapes. In \cite{zhang2018self}, unsupervised segmentation results are used to train a deep segmentation network on cystic lung regions, again in a slice-by-slice fashion. This approach might work well for certain organs, like the lungs, where an unsupervised technique can have good enough initial performance due to the good image contrast. However, completely unsupervised techniques might fail to generalize to organs where the boundary information is not as clear.
More recently, \cite{kervadec2019constrained}  introduced inequality constraints based on target-region size and image tags in the loss function of a CNN in order to train the network for weakly supervised segmentation.


\section{Method}
\label{sec:method}

In this work, we approach initial interactive segmentation using user-provided clicks on the extreme points of the organ of interest. The overall proposed algorithm for weakly supervised segmentation from extreme points can be divided into the following steps which are detailed below:
\begin{enumerate}
    \item Extreme point selection
    \item Initial segmentation from scribbles via random walker algorithm
    \item Segmentation via deep fully convolutional network
    \item Regularization using random walker algorithm
\end{enumerate}
Steps 2, 3, and 4 will be iterated until convergence. Here, convergence is defined based on the differences between two consecutive rounds of predictions as in \cite{zhang2018self}.

\paragraph{\textbf{1. Extreme point selection:}} 
Defining extreme points on the organ surface will allow the extraction of a bounding box around the organ (plus some padding $p$=20 mm in all our experiment). Bounding box selection significantly reduces the image content that the 3D FCN has to analyze and simplifies the machine learning problem, as previous work on cascaded approaches has shown \cite{roth2018spatial}.
Bounding boxes and extreme points on objects have been widely studied in the computer vision literature \cite{maninis2017deep}. Bounding boxes have a practical disadvantage in that the user often has to select the corners of bounding boxes that lie outside the object of interest. 
This is especially tricky to do for three-dimensional objects where the user typically has to navigate three multi-planar reformatted views (axial, coronal, sagittal) in order to achieve the task. Recent studies have also shown the time savings using extreme point selection brings for 2D object selection instead of traditional bounding box selection \cite{maninis2017deep,papadopoulos2017extreme}. 
At the same time, extreme points provide additional information to the segmentation model (which can be observed in our experimental section, Table \ref{table:results}. 
They lie on the object surface and we model them as an additional input channel together with the image intensities. This extra channel includes 3D Gaussians $G$ centered on each point location clicked by the user. This approach is similar to \cite{maninis2017deep} but here we extended it to 3D medical imaging problems.

\begin{figure*}[htbp!]
	\centering
	\begin{tabular}{cccc}
		\subfloat[(a)]{\adjincludegraphics[valign=c,height=1.4cm]{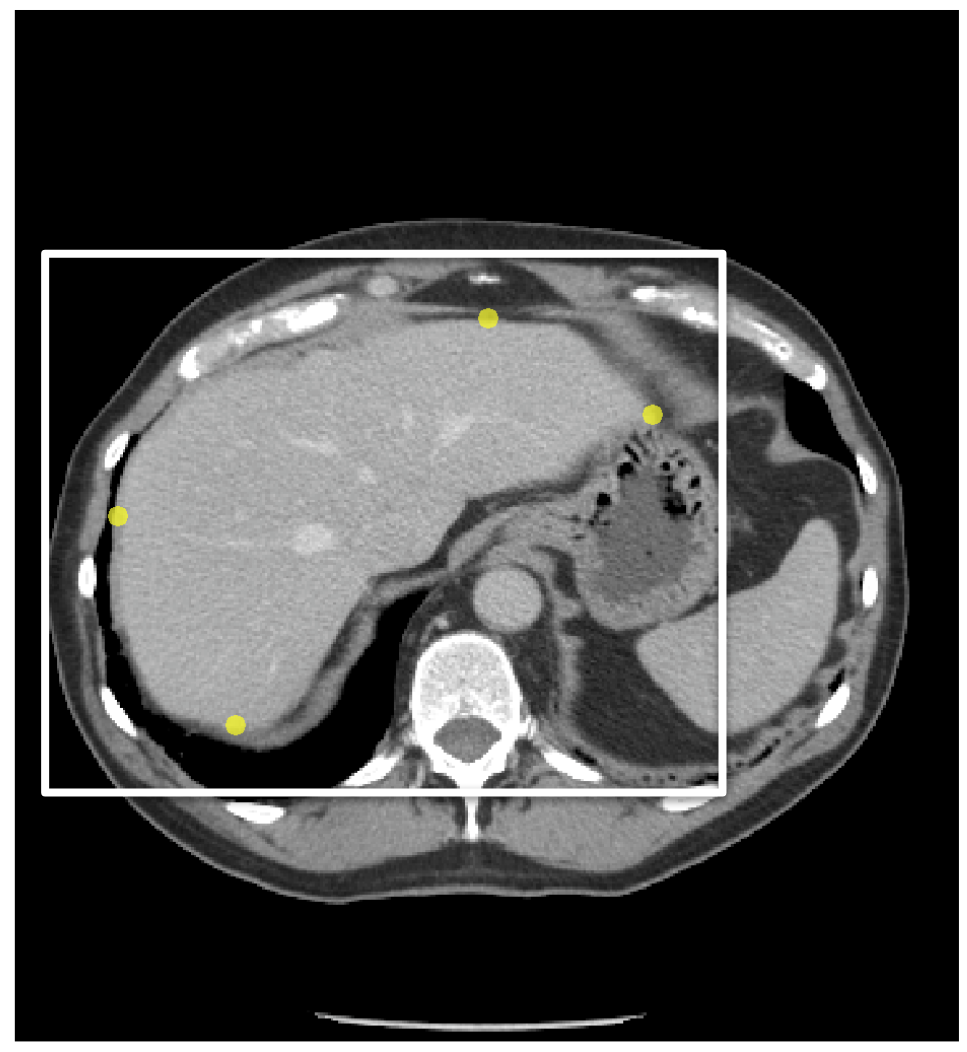}} &
		\hfill
		\subfloat[(b)]{\adjincludegraphics[valign=c,height=1.4cm]{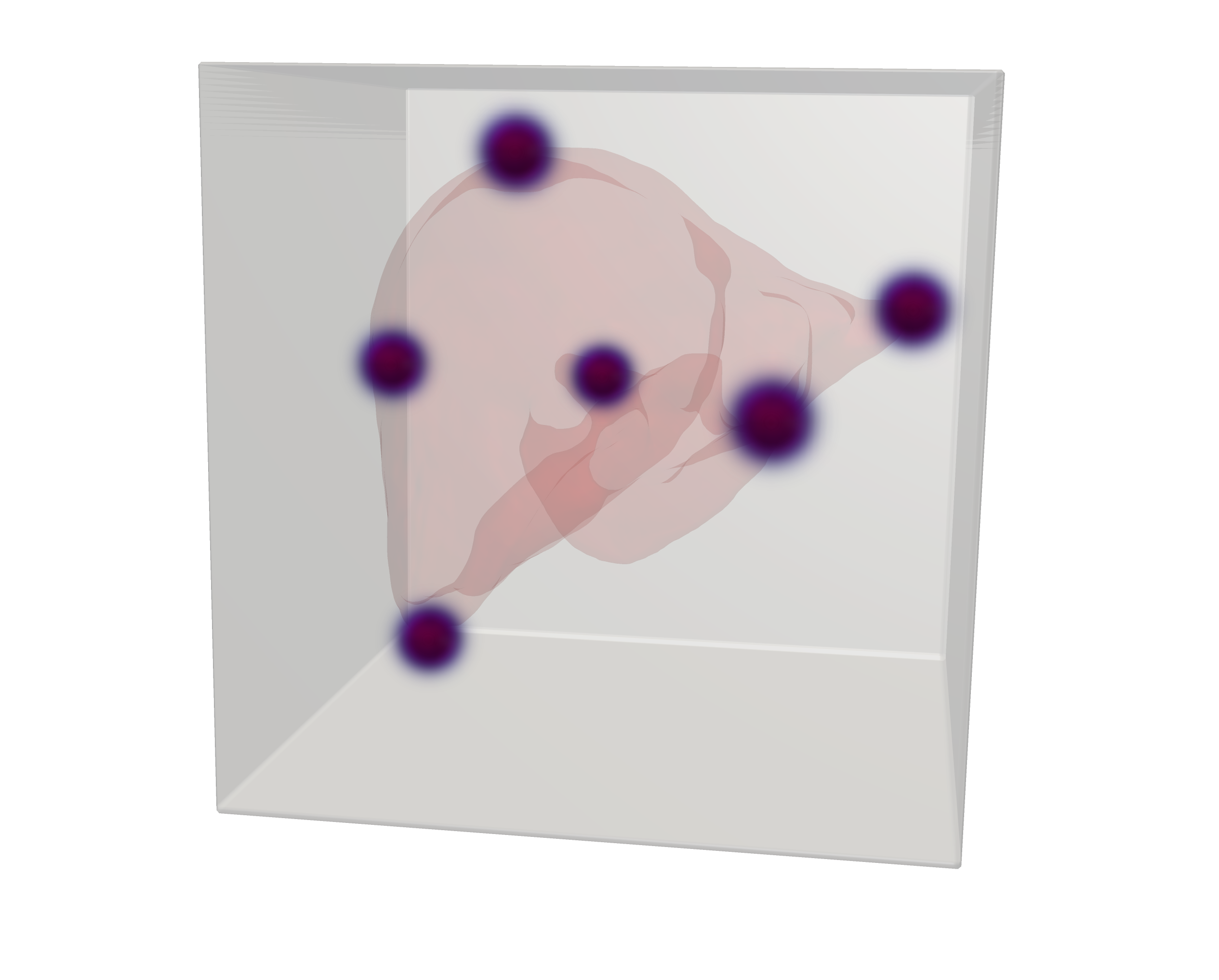}} &  
		\hfill
		\subfloat[(c)]{\adjincludegraphics[valign=c,height=1.0cm]{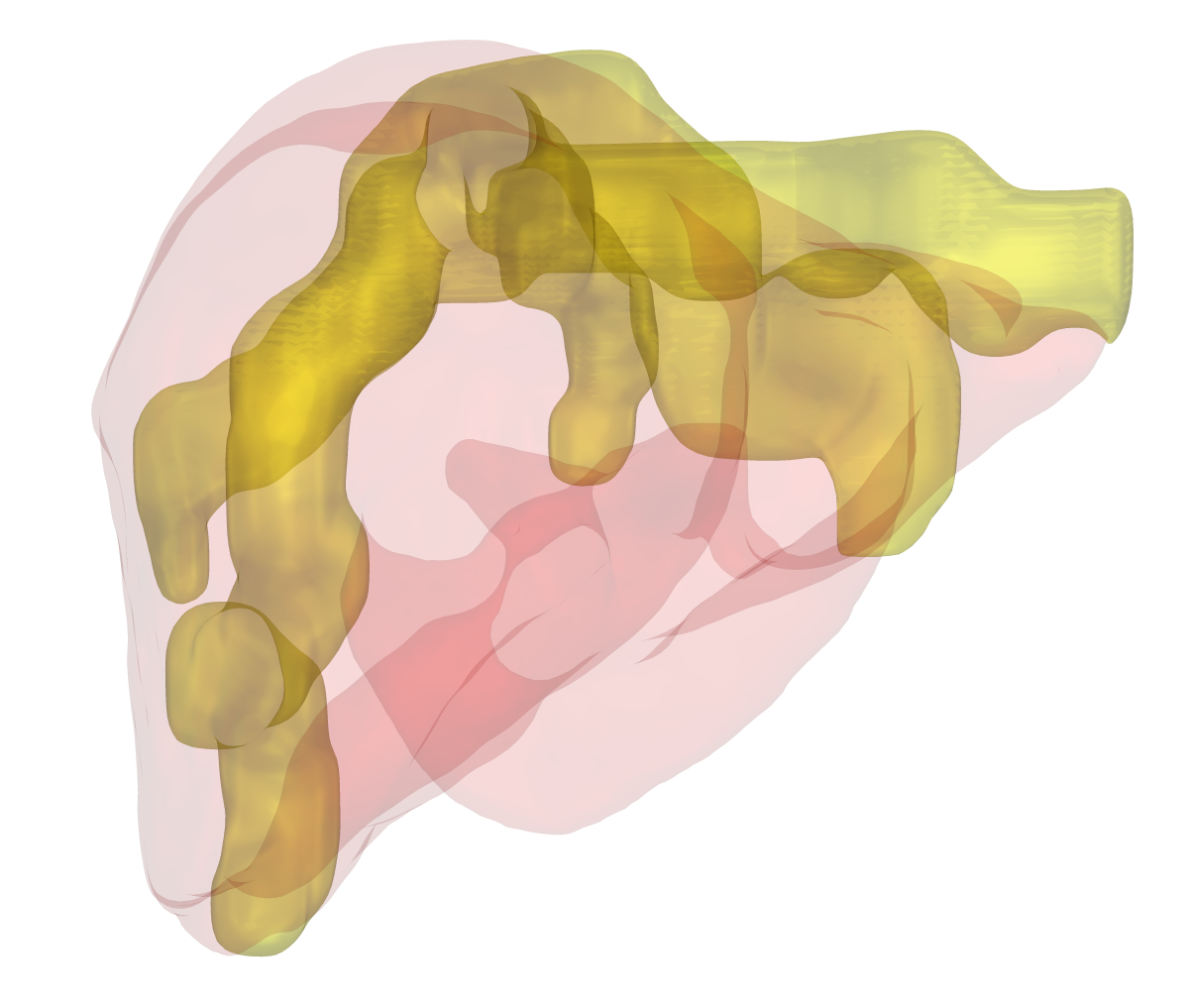}} &		
		\hfill
		\subfloat[(d)]{\adjincludegraphics[valign=c,height=1.4cm]{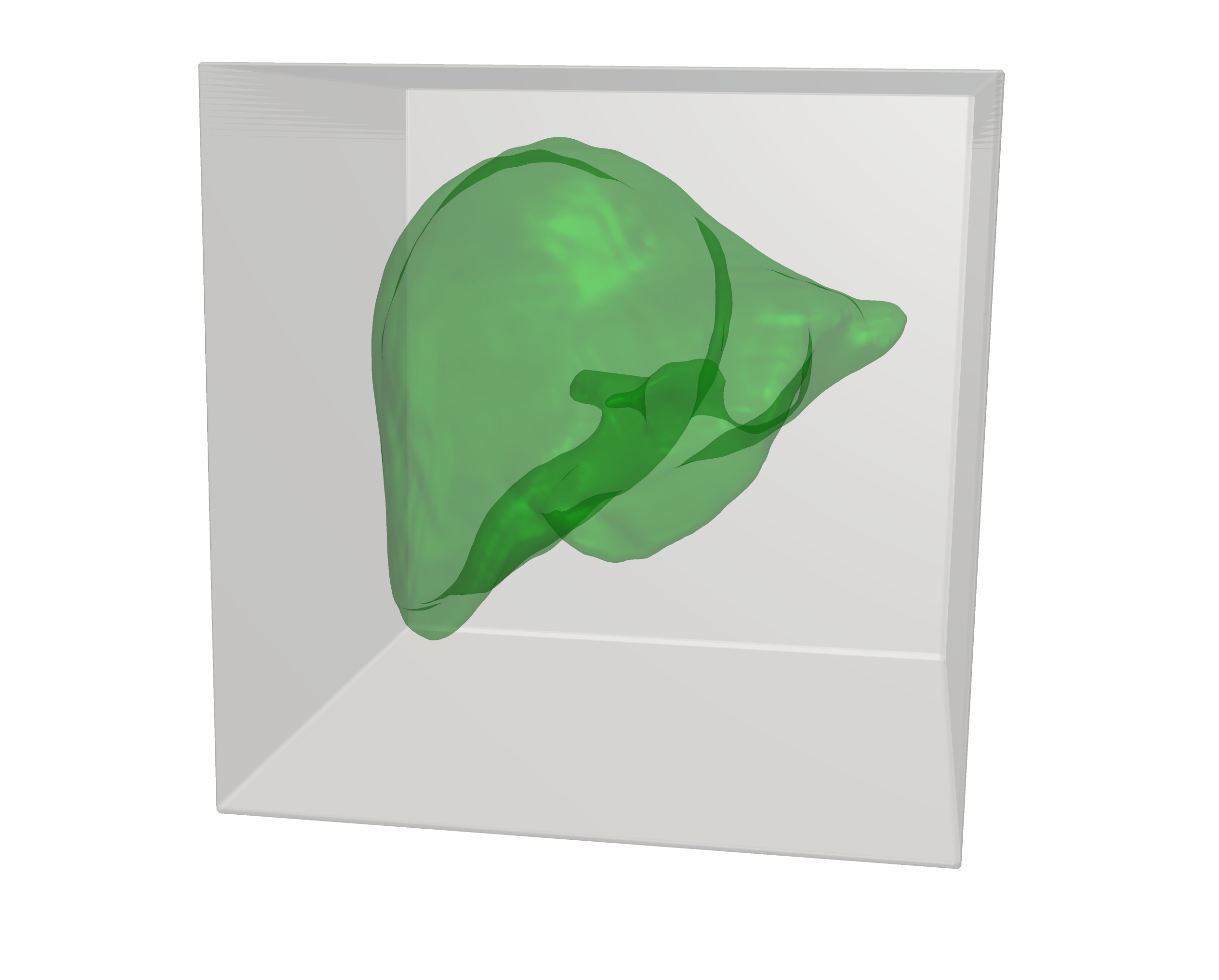}}
	\end{tabular}
	\caption{Our weakly supervised segmentation framework. (a) The user selects extreme points that define the organ of interest (here the liver) in 3D space. (b) Extreme points are modeled as Gaussians in an extra image channel which is fed to a 3D segmentation model. (c) Foreground scribbles are generated automatically to initialize random walker (the ground truth surface is shown in red for reference). (d) Model returns the segmentation results.
	\label{fig:dextr3D}}
\end{figure*}

Figure \ref{fig:dextr3D} illustrates our approach. We ask the user to click on six extreme points (here four are shown in axial view) that describe the largest extent of the organ. These points are then used to compute a bounding box $B$ automatically, including some padding $p$.

\paragraph{\textbf{2. Initial segmentation from scribbles via random walker algorithm:}}
In order to make use of extreme point clicks as a weak supervision signal, we turn them into a probability map $\hat{Y}$ than can act as a pseudo dense label map for driving a 3D FCN to learn the segmentation task. 
Based on the initial set of extreme points, we compute a set of foreground and background scribbles that act as the input seeds for the random walker algorithm \cite{grady2006random}. We compute Dijkstra's shortest path \cite{dijkstra1959note} between each extreme point pair along each image dimension, where we model the distance between neighboring voxels by their gradient magnitude {\scriptsize $D = \sqrt{ \left(\frac{\partial f}{\partial x}\right)^2 + \left(\frac{\partial f}{\partial y}\right)^2 + \left(\frac{\partial f}{\partial z}\right)^2  }$}.
Here, the shortest path result can be seen as an approximation of the geodesic distance \cite{wang2018deepigeos} between the two extreme points in each dimension. Figure \ref{fig:dextr3D} shows the foreground scribbles used as input seeds to the random walker algorithm. In order to increase the number of foreground seeds, each path is also dilated with a 3D ball structure element of $r_\mathrm{foreground}=2$. The background seeds are defined as the dilated and inverted version of the input scribbles. While the amount of dilation does depend on the size of the organ of interest, we typically dilate with a ball structure element of radius $r_\mathrm{background}=30$ which achieves good initial seeds for organs like spleen, and liver.


Next, the random walker algorithm \cite{grady2006random} is used to generate an initial prediction map $\hat{Y}$ based on the background $s_0$ and foreground $s_1$ scribbles described above. 
The random walker basically solves the diffusion equation between voxels defined as source and sink as defined by the scribbles $S$. Here, the 3D volume is defined as a graph $G(E,V)$ with edges $e \in E$ and vertices $v \in V$. The edge between two vertices $v_i$ and $v_j$ is denoted as $e_{ij}$ and can be assigned a weight $w_{ij}$ based on the image intensities gradients. Furthermore, the degree of a given vertex is defined by $d_i = \sum{ w_{ij} }$.
We solve the diffusion equation in order to get a probability $p(\omega|x) = x^\omega_j$ for each vertex $v_i$ to belong to the foreground class $\omega_1$. Here, $L$ is the Laplacian of the weighted image graph $G$ with each element of the matrix defined as:
\begin{equation}
    L_{ij} = 
    \begin{cases}
        d_i,& \text{if } i = j,\\
        -w_{ij},& \text{if } i \text{ and } j \text{ are adjacent voxels},\\
        0,& \text{otherwise}
    \end{cases}
\end{equation}
The weights between adjacent voxels are defined as $w_{ij}=e^{-\beta|z_j-z_i|^2}$ to make diffusion between similar voxel intensities $z_i$ and $z_j$ easier. While $\beta$ is a tunable hyperparameter that controls the amount of diffusion, we keep it fixed at $\beta=130$ in all our experiments.



\paragraph{\textbf{3. Segmentation via deep fully convolutional network:}}
Next, given all pairs of images $X$ and pseudo labels $\hat{Y}$, we can train a fully convolutional neural network to segment the given foreground class, with $P(X) = f(X)$. 
Our network architecture of choice follows the encoder-decoder network proposed in \cite{liu20183d}, utilizing an-isotropic ($3 \times 3 \times 1$) kernels in the encoder path in order to make use of pretrained weights from 2D computer vision tasks. As in \cite{liu20183d}, we initialize from \textit{ImageNet} pretrained weights using a ResNet-18 encoder branch. 
While the initial weights are learned from 2D, all convolutions are still applied in a full 3D fashion throughout the network, allowing it to efficiently learn 3D features from the image.
The Dice loss \cite{milletari2016v} has been established as the objective function of choice for medical image segmentation tasks. Its properties allow automatic scaling to unbalanced labeling problems. At the same time, it also naturally adapts to the comparing probability maps without any modifications to the original formulation:
\begin{equation}
    \mathcal{L}_{Dice} = 1 - \frac{2\sum_{i=1}^{N}y_i\hat{y}_i}{\sum_{i=1}^{N} y_i^2 + \sum_{i=1}^{N} \hat{y}_i^2}
\end{equation}
Here, $y_i$ is the predicted probability from our network $f$ and $\hat{y_i}$ is the weak label probability from our pseudo label map $\hat{Y}$ at voxel $i$.

\paragraph{\textbf{4. Regularization using random walker algorithm:}}
We could stop our learning after the segmentation network $f$ above is trained on the pseudo labels $\hat{Y}$. However, we notice that an additional regularization step by an additional random walker segmentation as described above can be very beneficial to the convergence of our weakly-supervised segmentation approach. This finding is similar in spirit to \cite{rajchl2017deepcut}, where a DenseCRF is utilized after each round of CNN training in order to introduce regularization to the segmentation output.
In order to increase the amount of regularization the random walker can bring to the network's predictions, we add an area of uncertainty by eroding the foreground prediction $P(X)>=0.5$ and eroding the background $P(X)<0.5$ both with a ball structure element of radius $r_\mathrm{randomwalker}=4$ in all our experiments. This allows the random walker to produce new predictions around the boundary of the foreground object that differ from the previous 3D FCN predictions and in turn, help the next iteration to learn new features from the same set of training images, and not to get stuck in a local optimum. In fact, we notice that without this step, our weakly supervised segmentation framework becomes unstable and does not easily converge to a satisfying performance.

\section{Experiments \& Results}
\label{sec:experiments}
\begin{figure*}[htbp]
	\centering
	\begin{tabular}{cccc}
		\subfloat{\adjincludegraphics[valign=c,height=1.2cm]{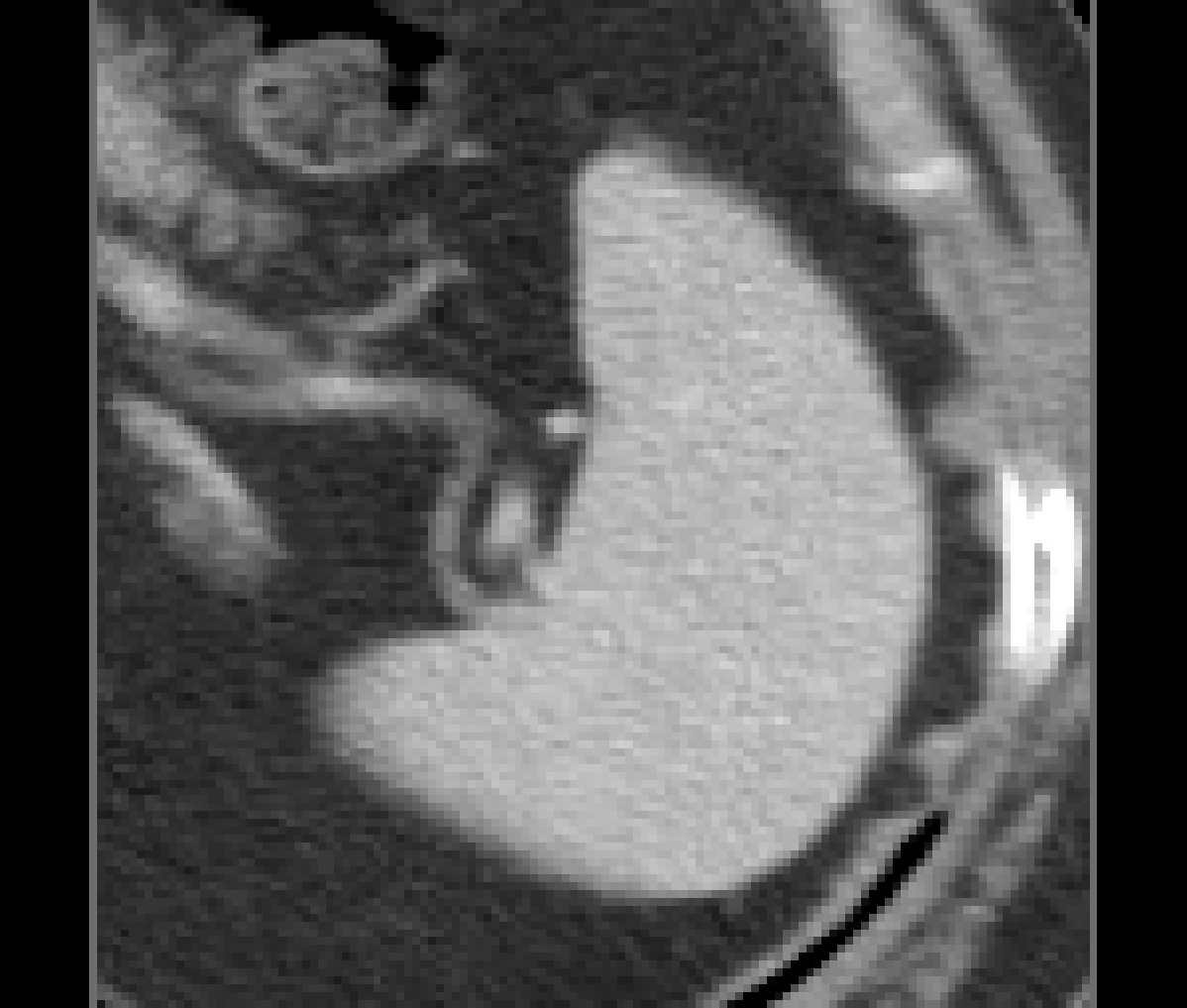}} &
		\hfill
		\subfloat{\adjincludegraphics[valign=c,height=1.2cm]{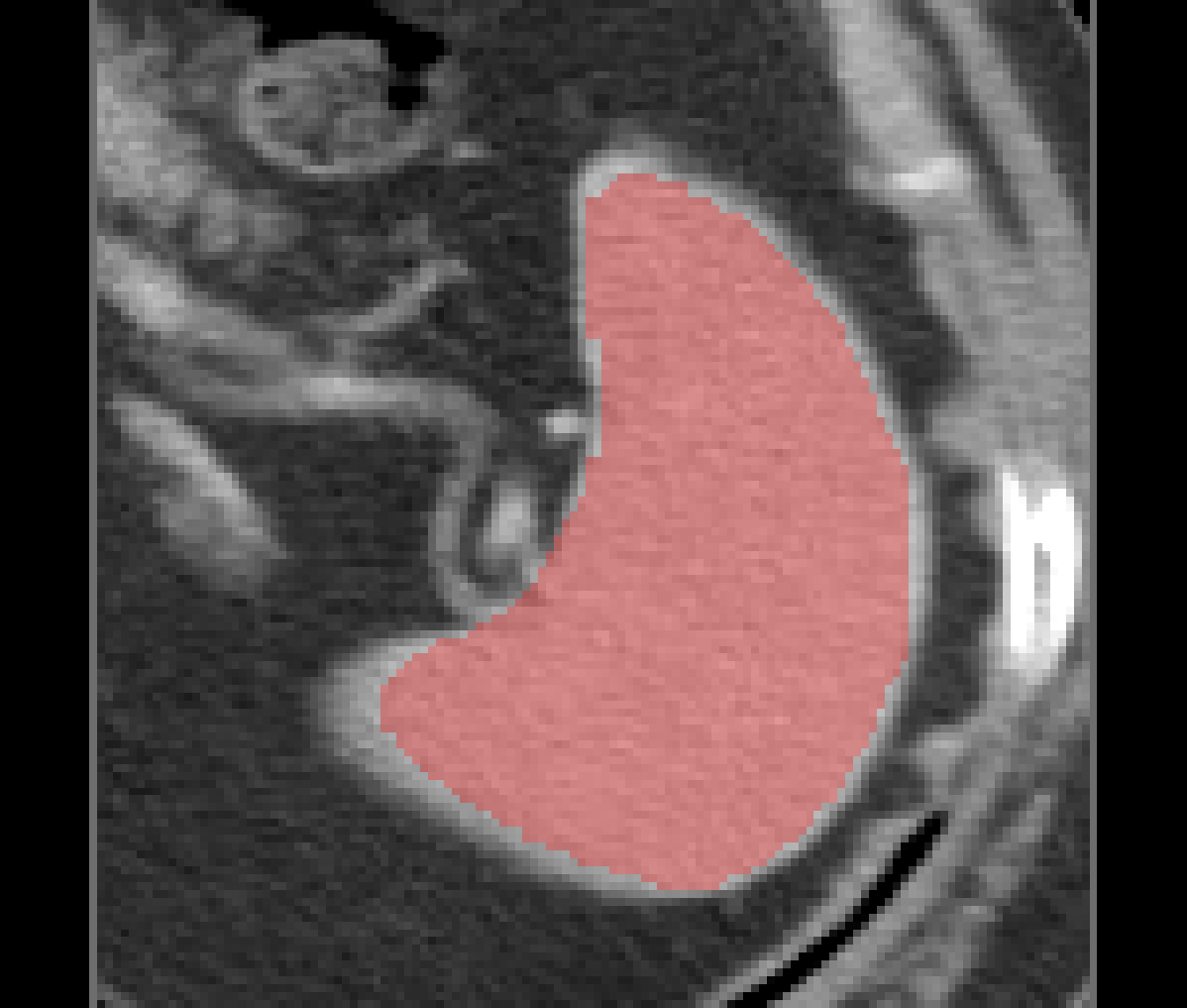}} & 
		\hfill
		\subfloat{\adjincludegraphics[valign=c,height=1.2cm]{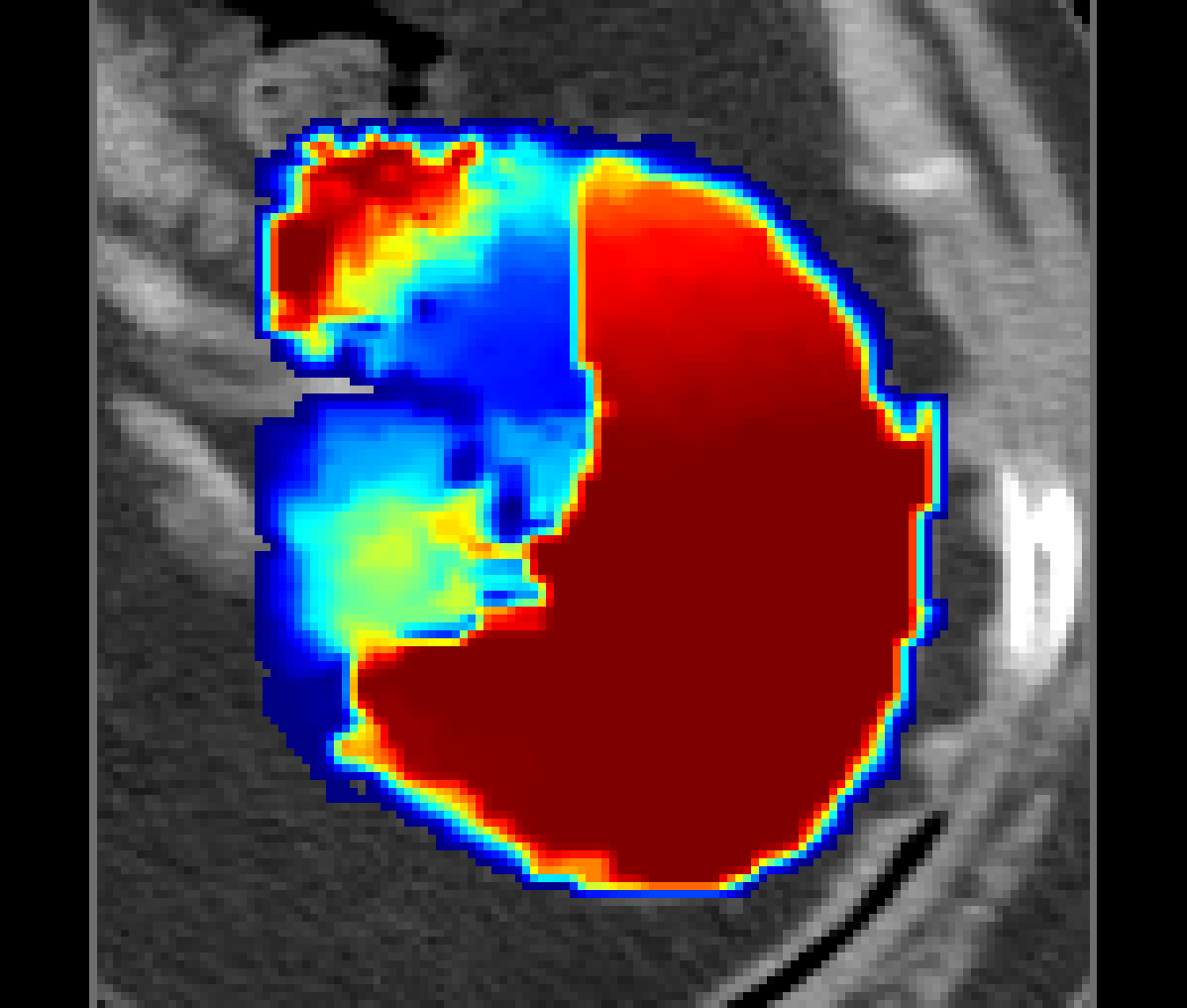}} &
		\hfill
		\subfloat{\adjincludegraphics[valign=c,height=1.2cm]{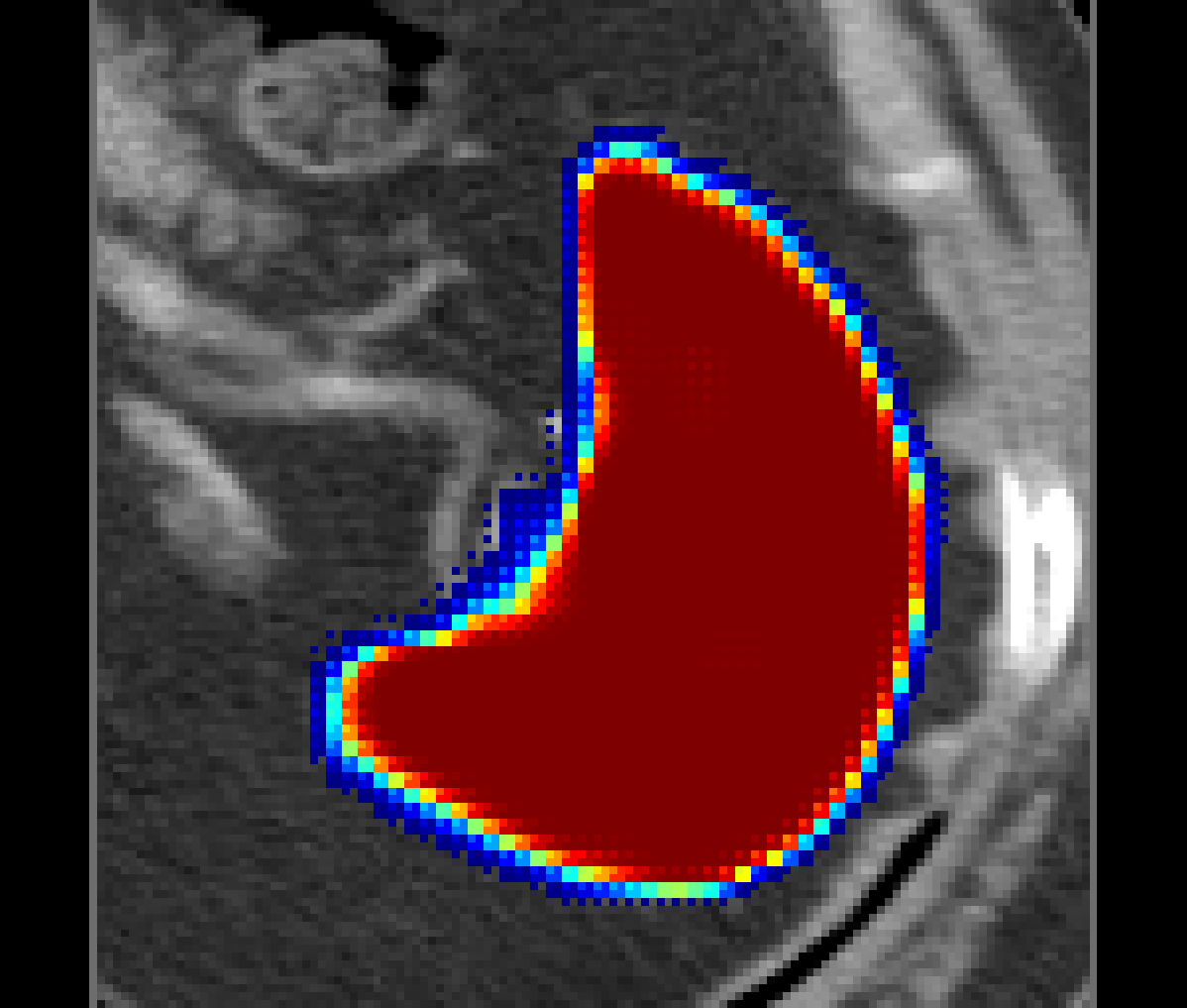}} \\
		\subfloat{\adjincludegraphics[valign=c,height=1.2cm]{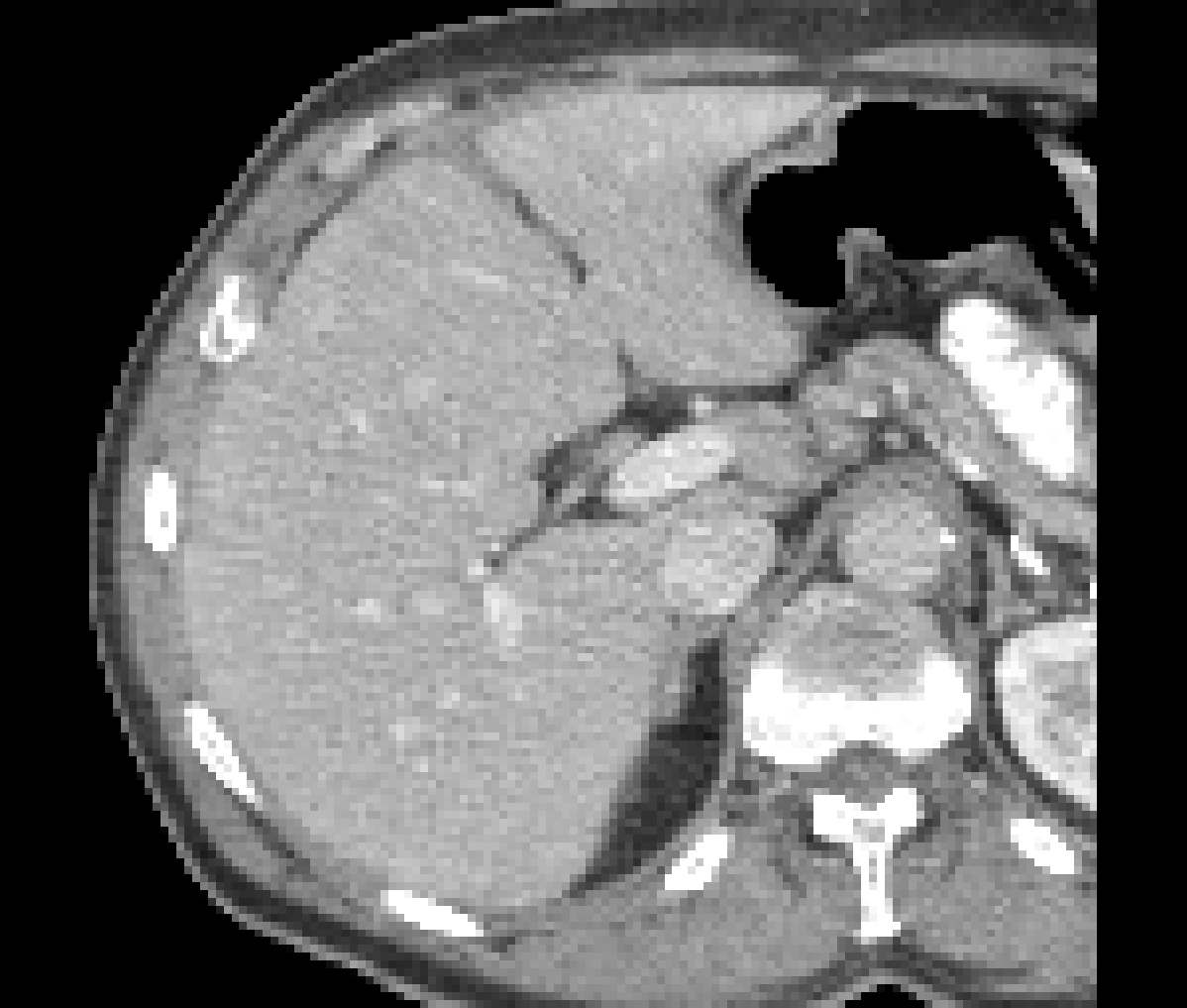}} &
		\hfill
		\subfloat{\adjincludegraphics[valign=c,height=1.2cm]{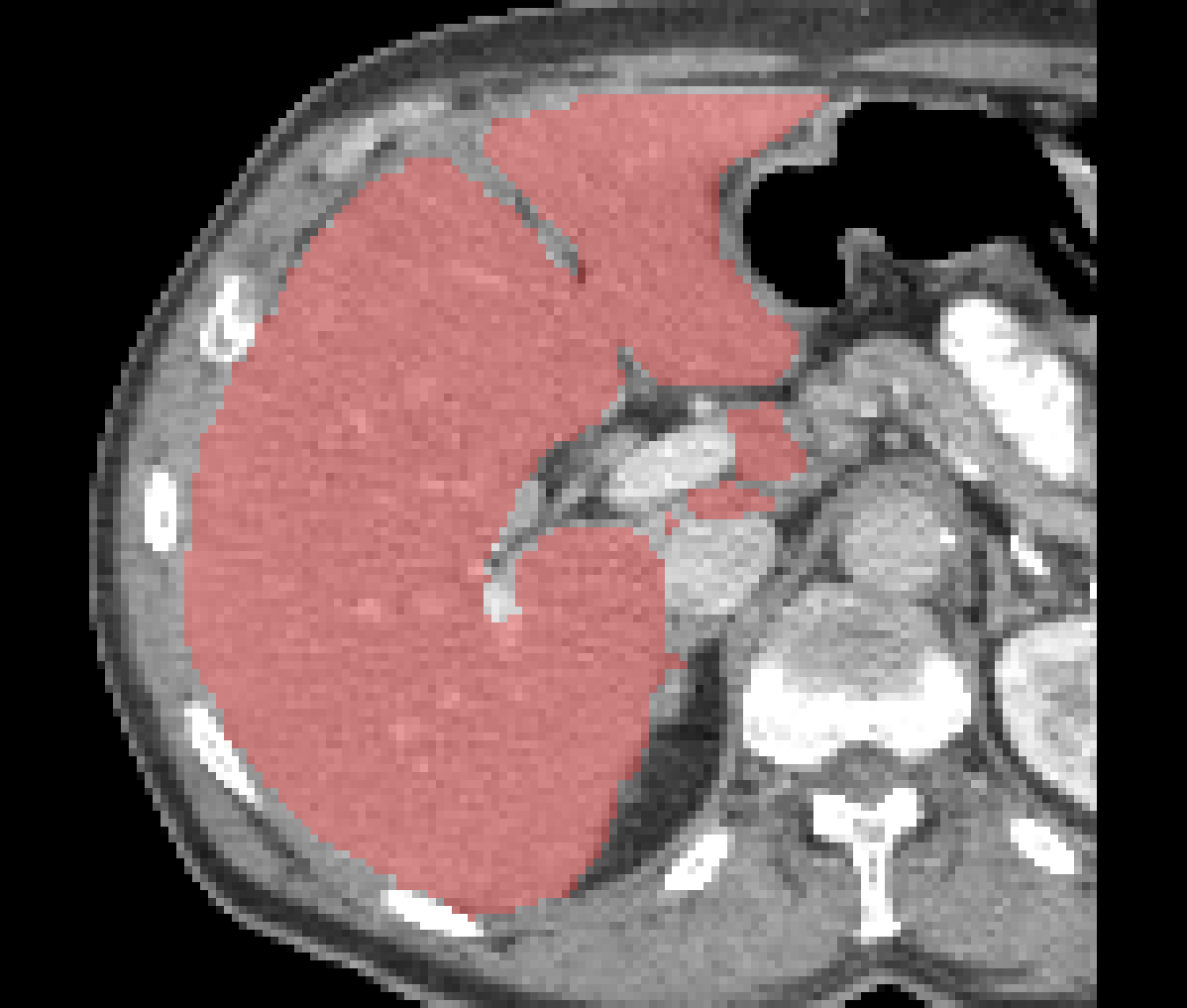}} & 
		\hfill
		\subfloat{\adjincludegraphics[valign=c,height=1.2cm]{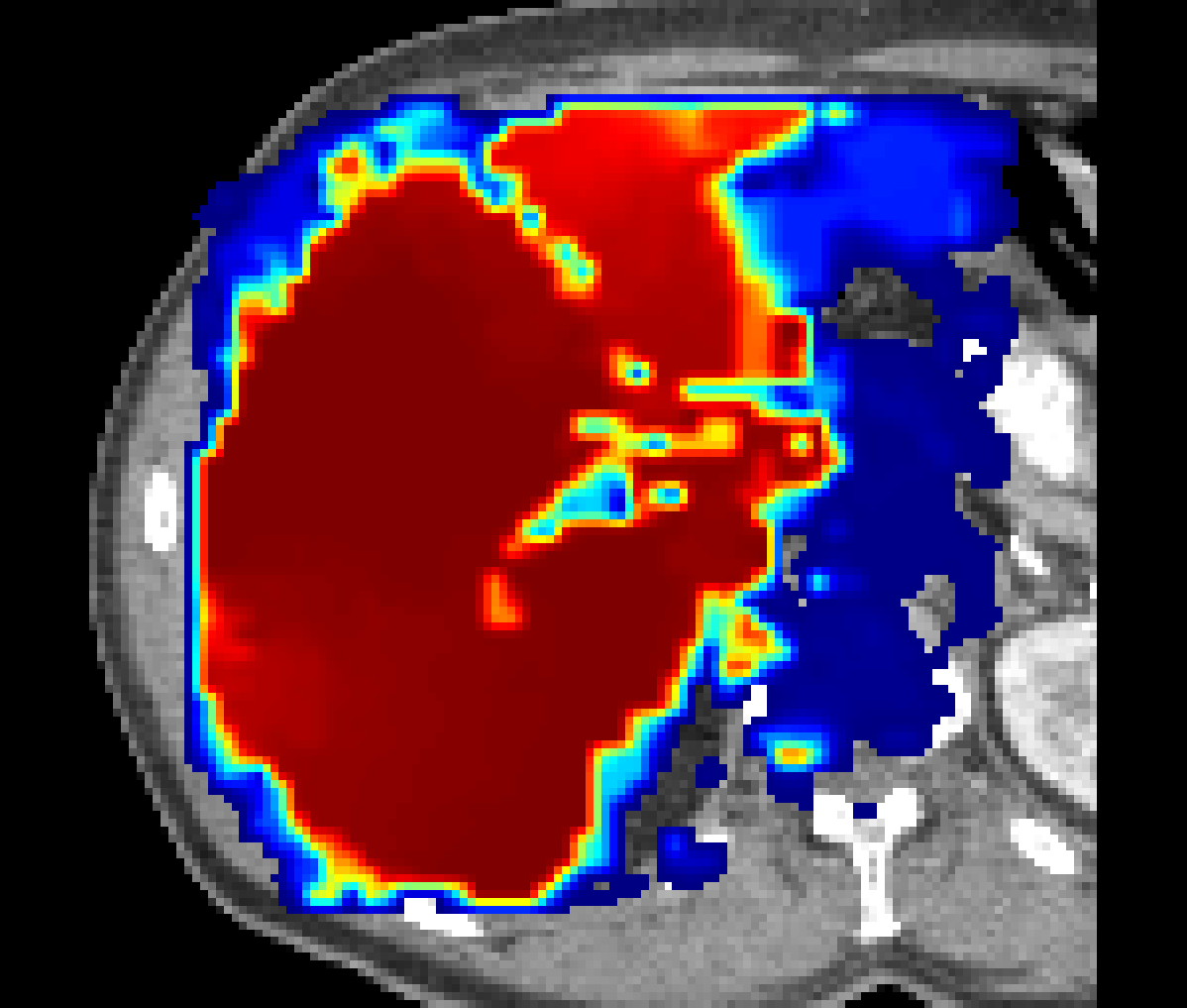}} &
		\hfill
		\subfloat{\adjincludegraphics[valign=c,height=1.2cm]{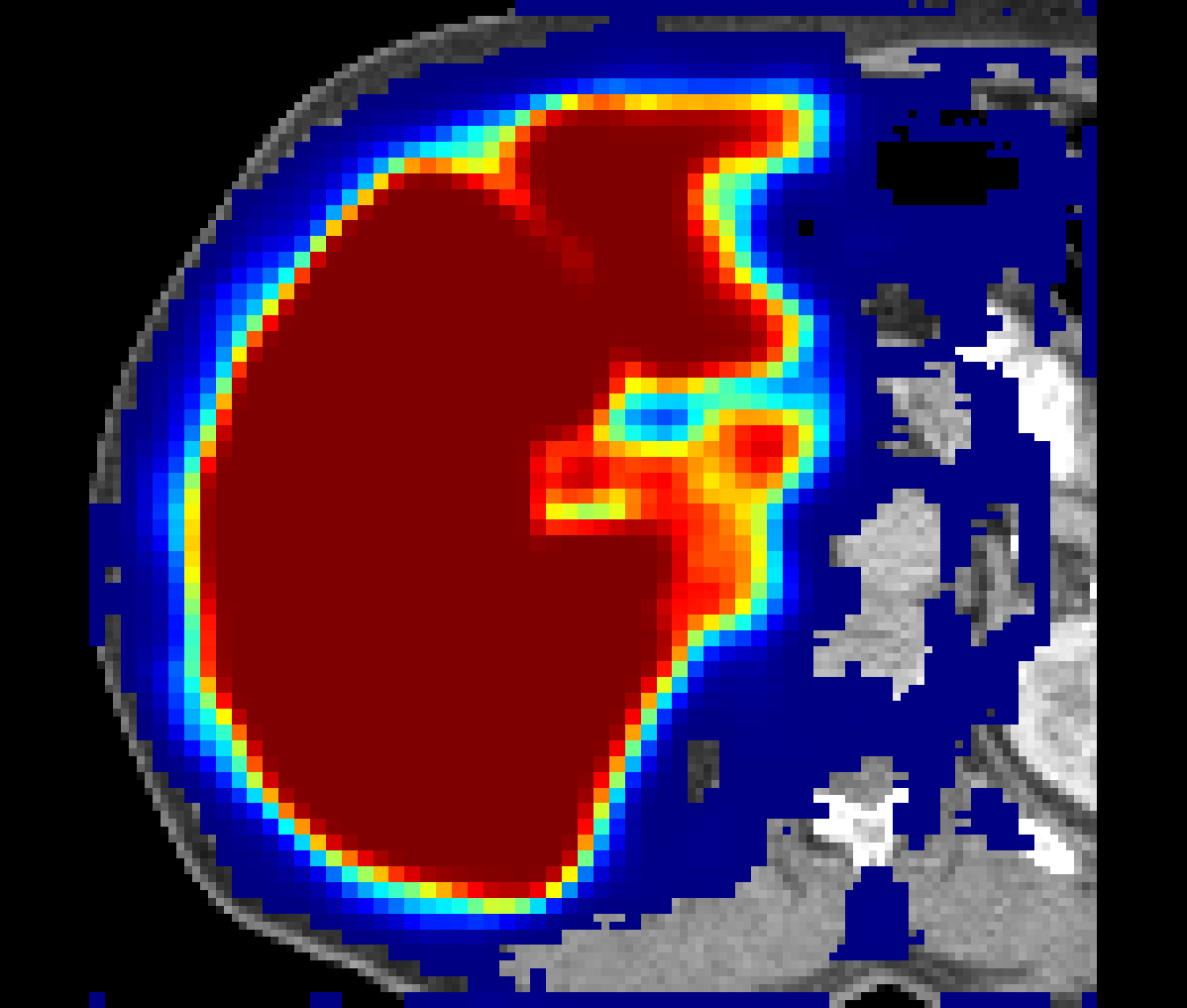}} \\
		\subfloat{\adjincludegraphics[valign=c,height=1.2cm]{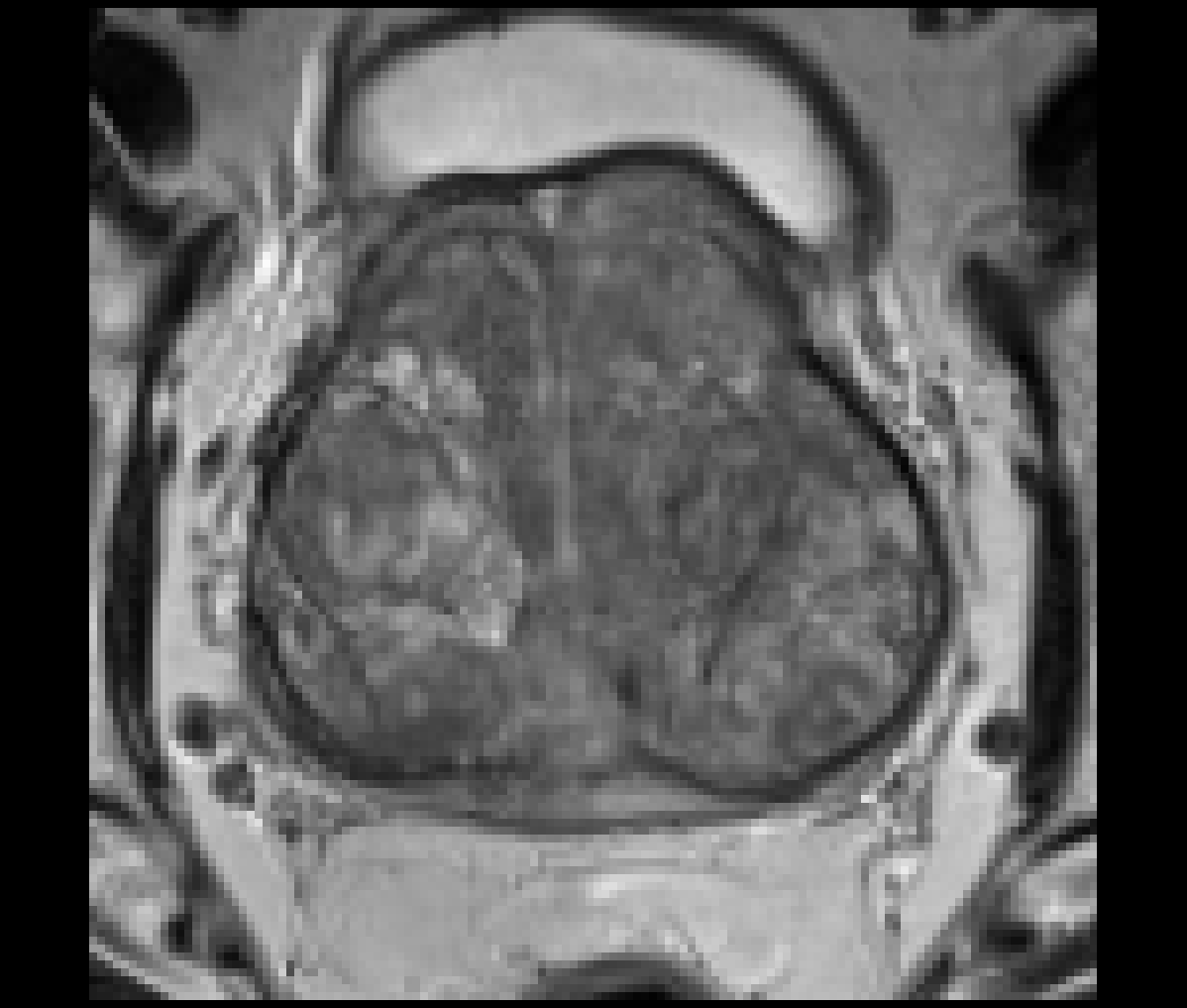}} &
		\hfill
		\subfloat{\adjincludegraphics[valign=c,height=1.2cm]{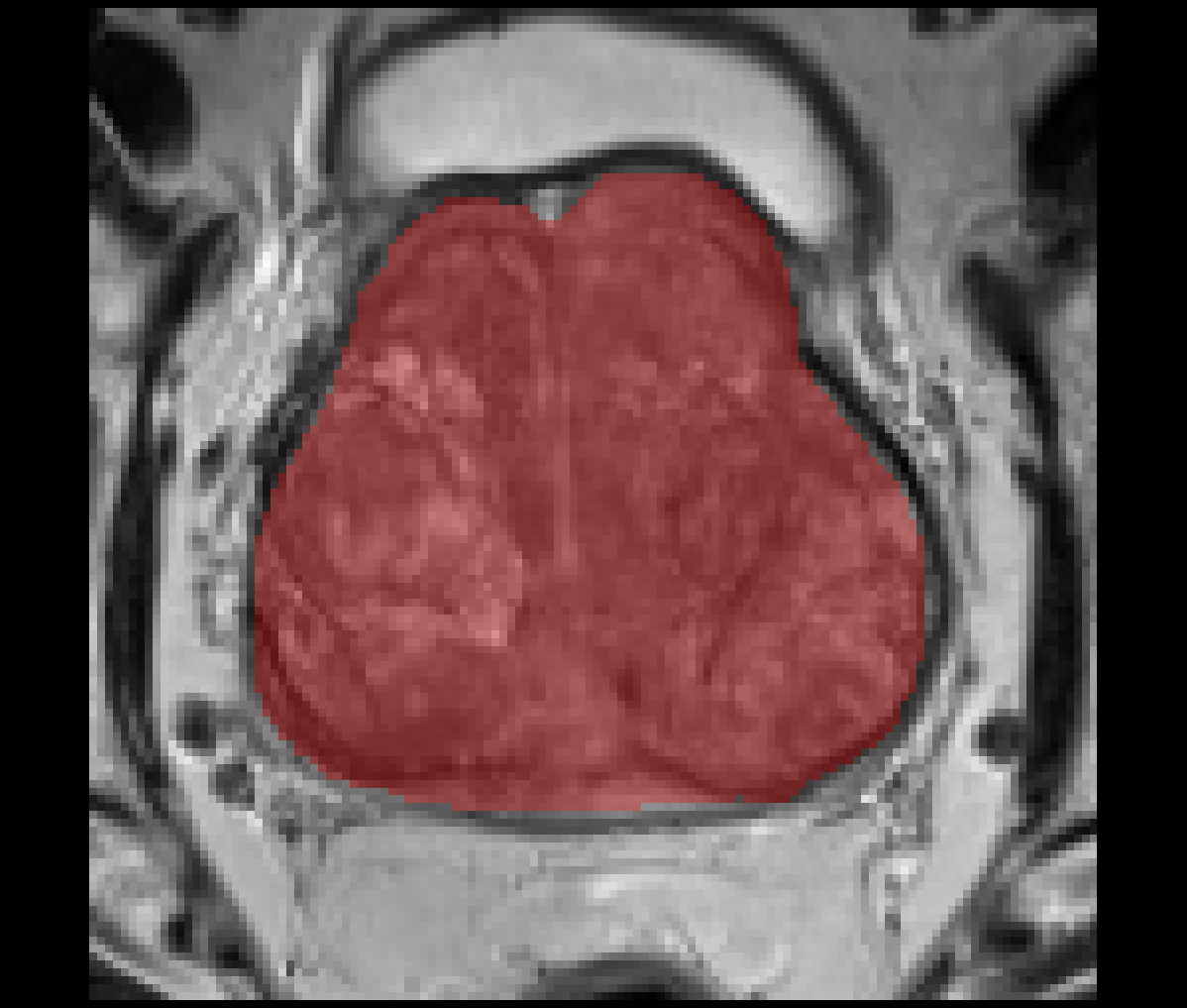}} & 
		\hfill
		\subfloat{\adjincludegraphics[valign=c,height=1.2cm]{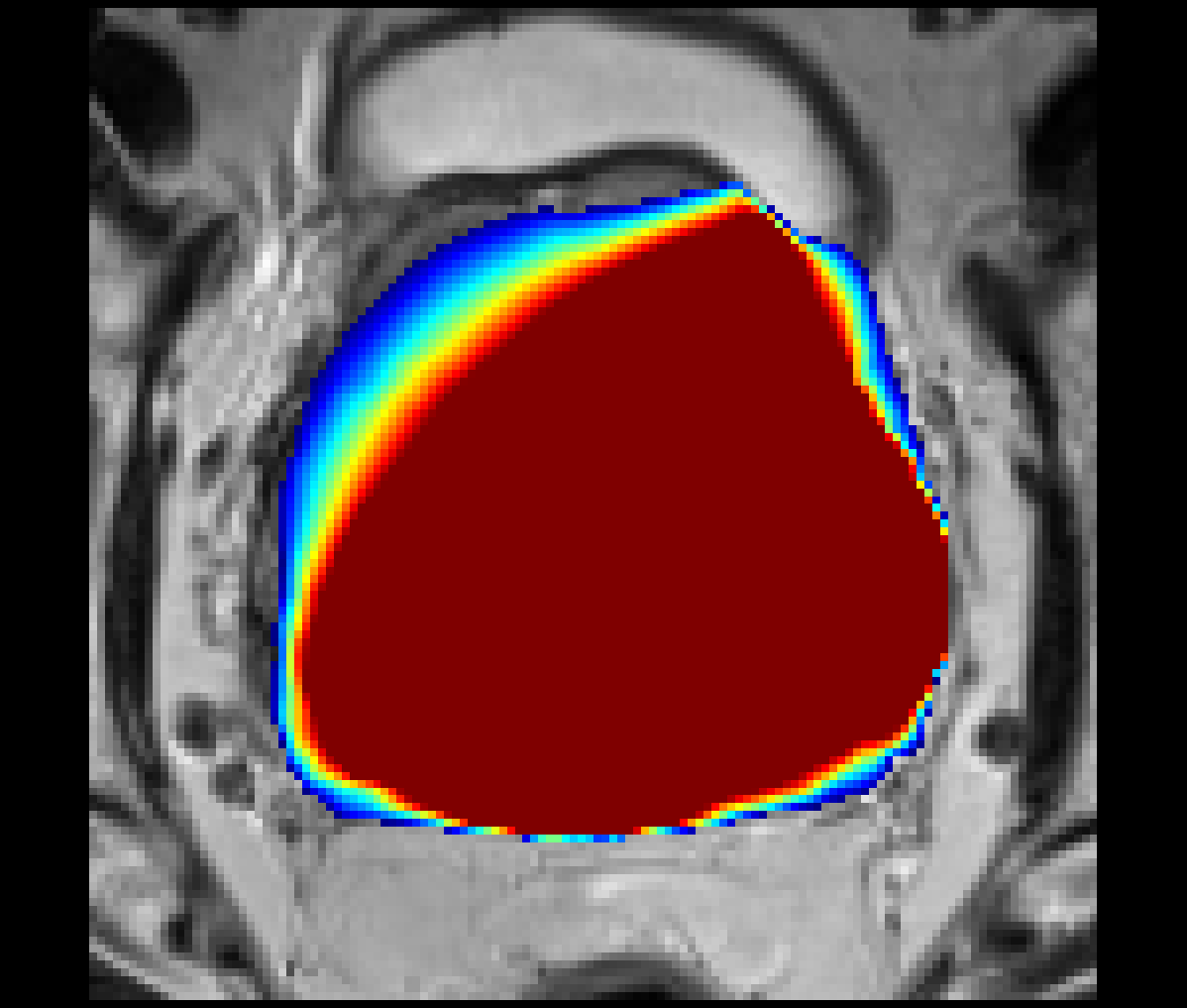}} &
		\hfill
		\subfloat{\adjincludegraphics[valign=c,height=1.2cm]{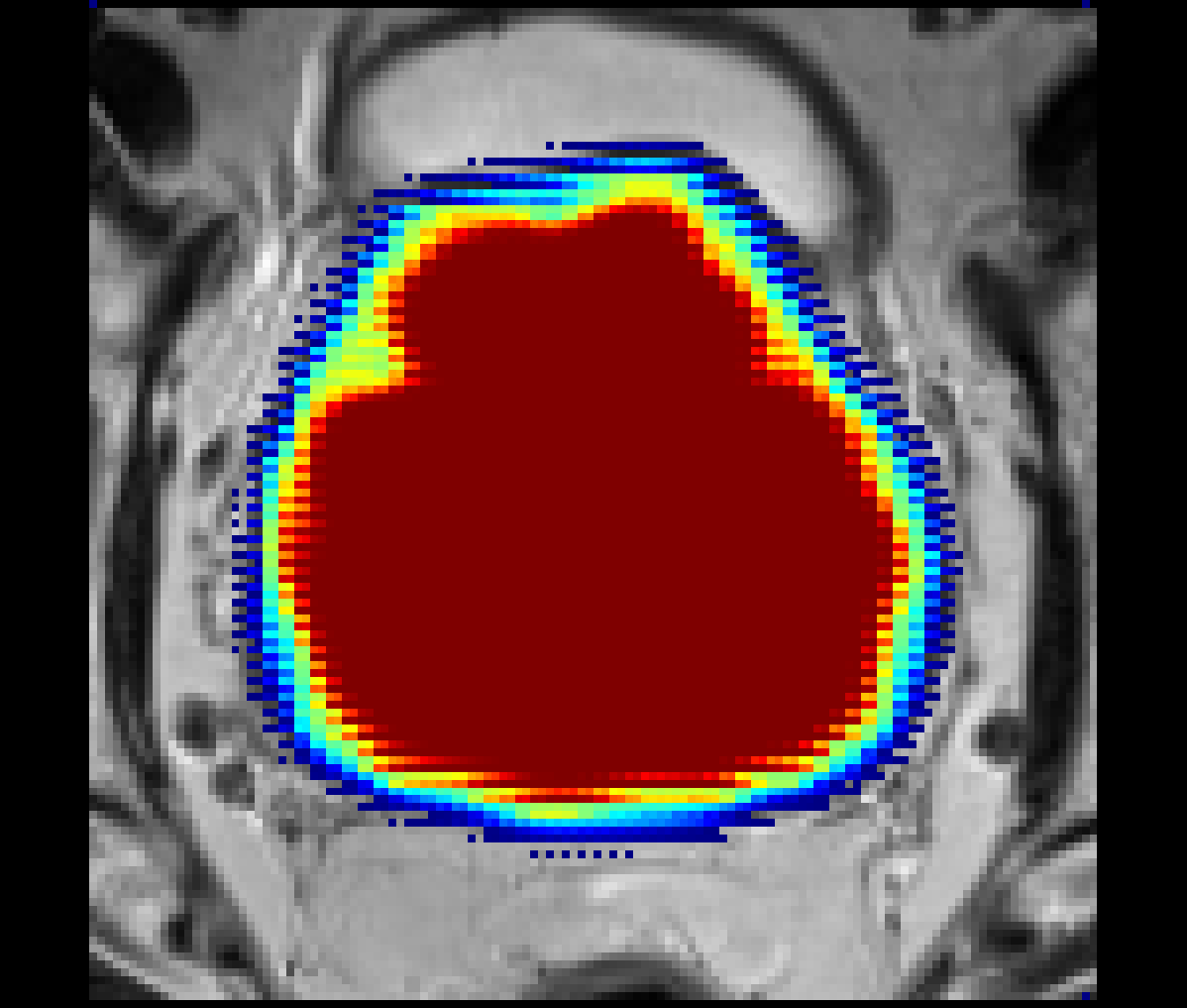}} \\
		\subfloat[(a)]{\adjincludegraphics[valign=c,height=1.2cm]{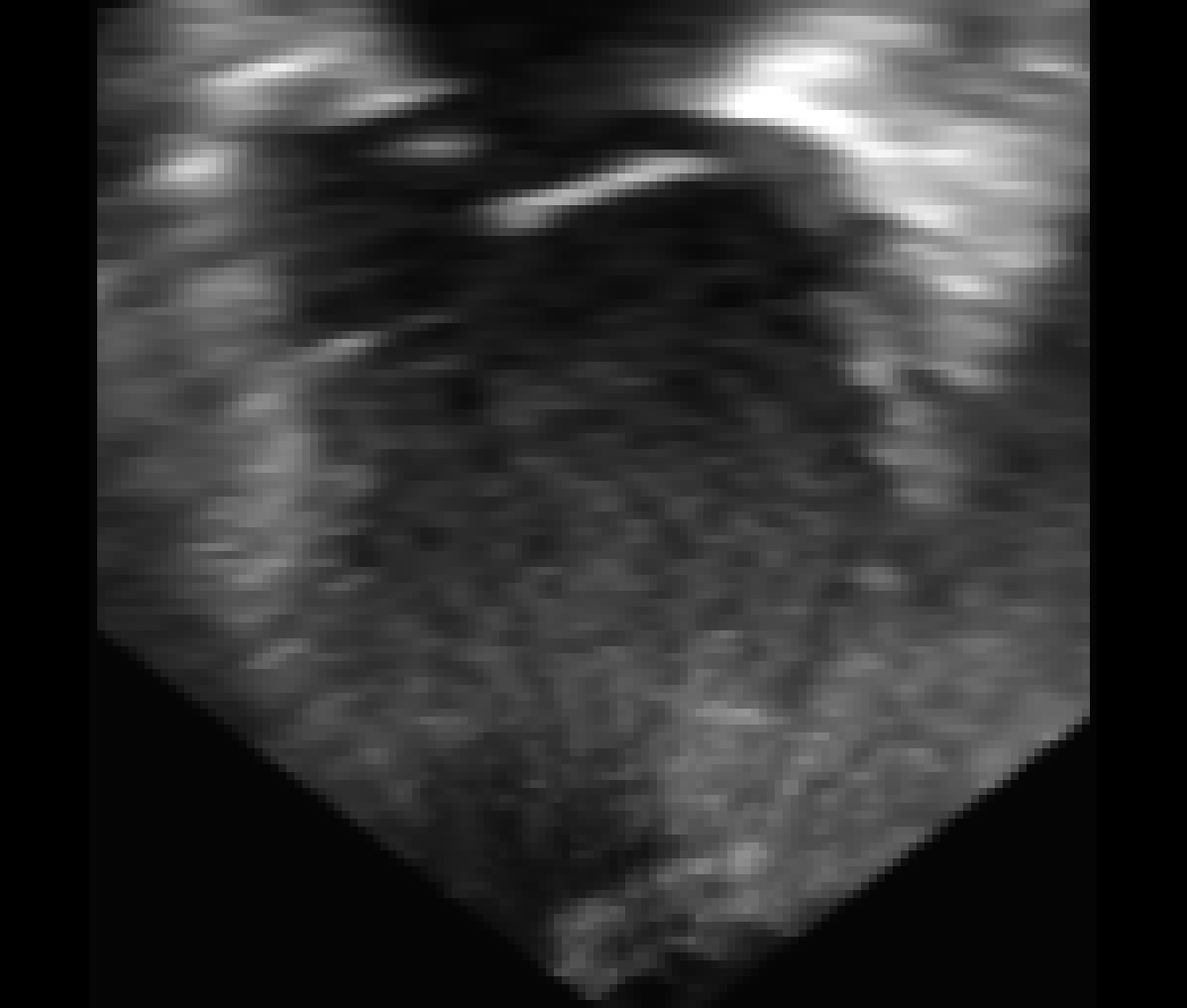}} &
		\hfill
		\subfloat[(b)]{\adjincludegraphics[valign=c,height=1.2cm]{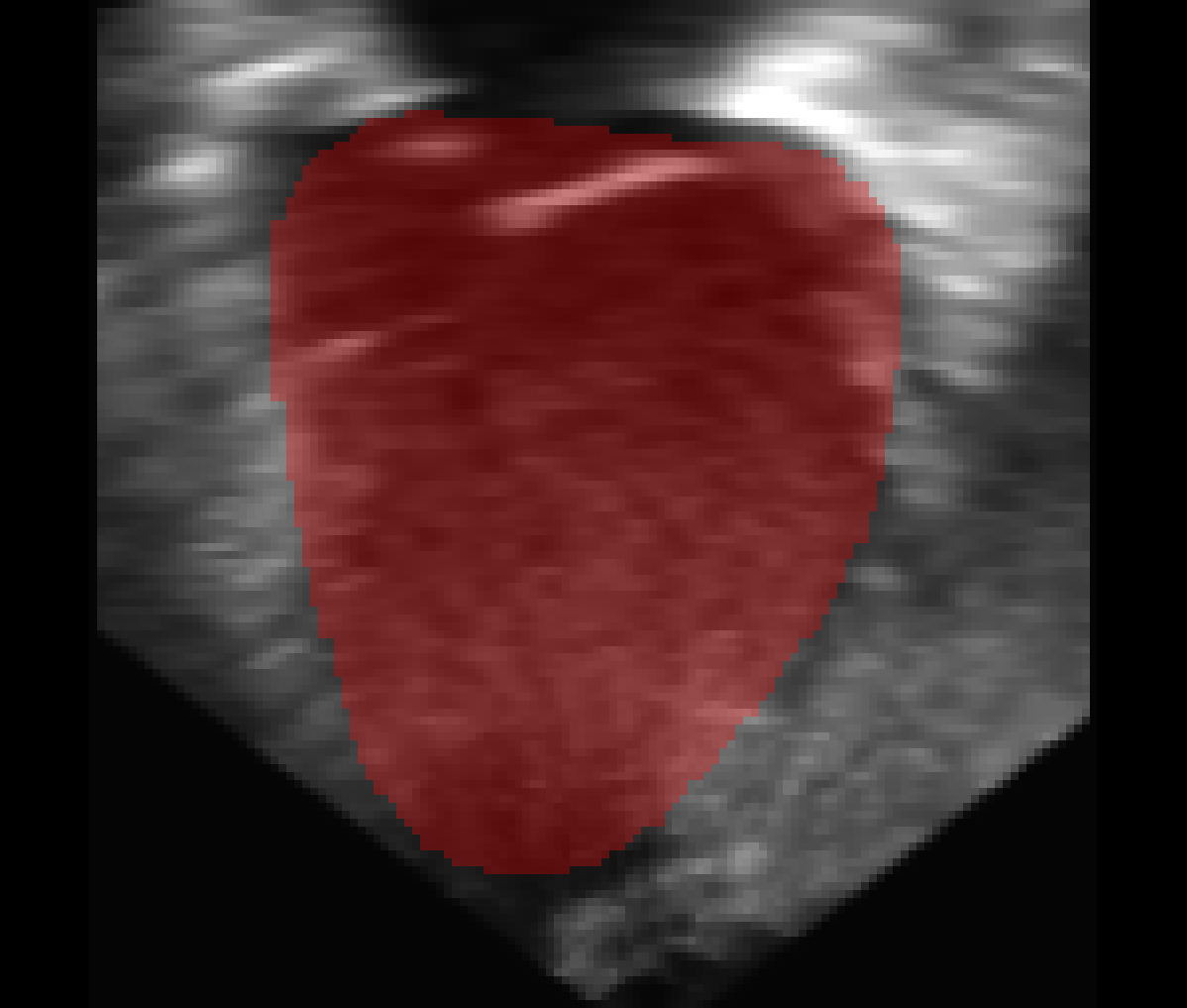}} & 
		\hfill
		\subfloat[(c)]{\adjincludegraphics[valign=c,height=1.2cm]{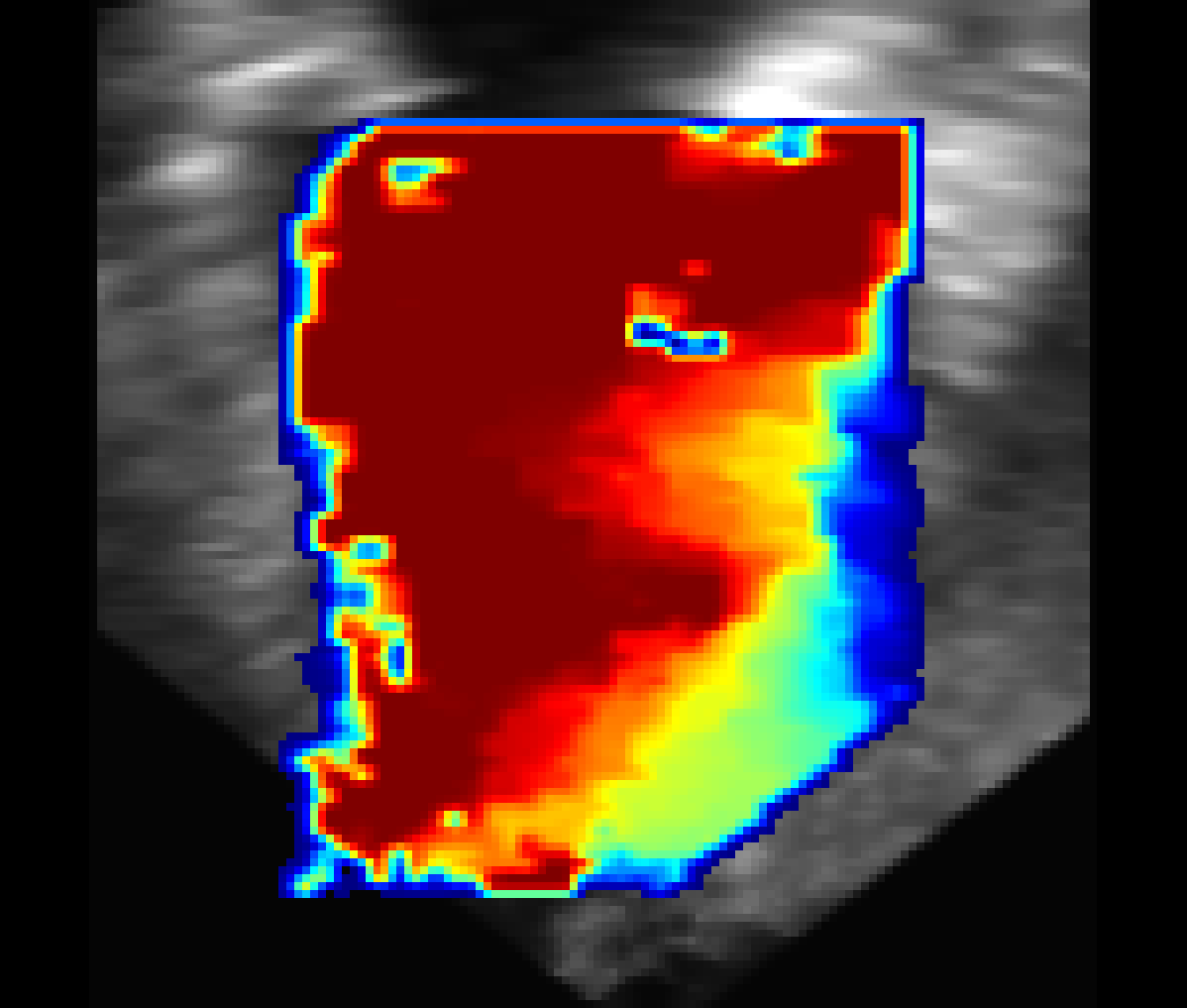}} &
		\hfill
		\subfloat[(d)]{\adjincludegraphics[valign=c,height=1.2cm]{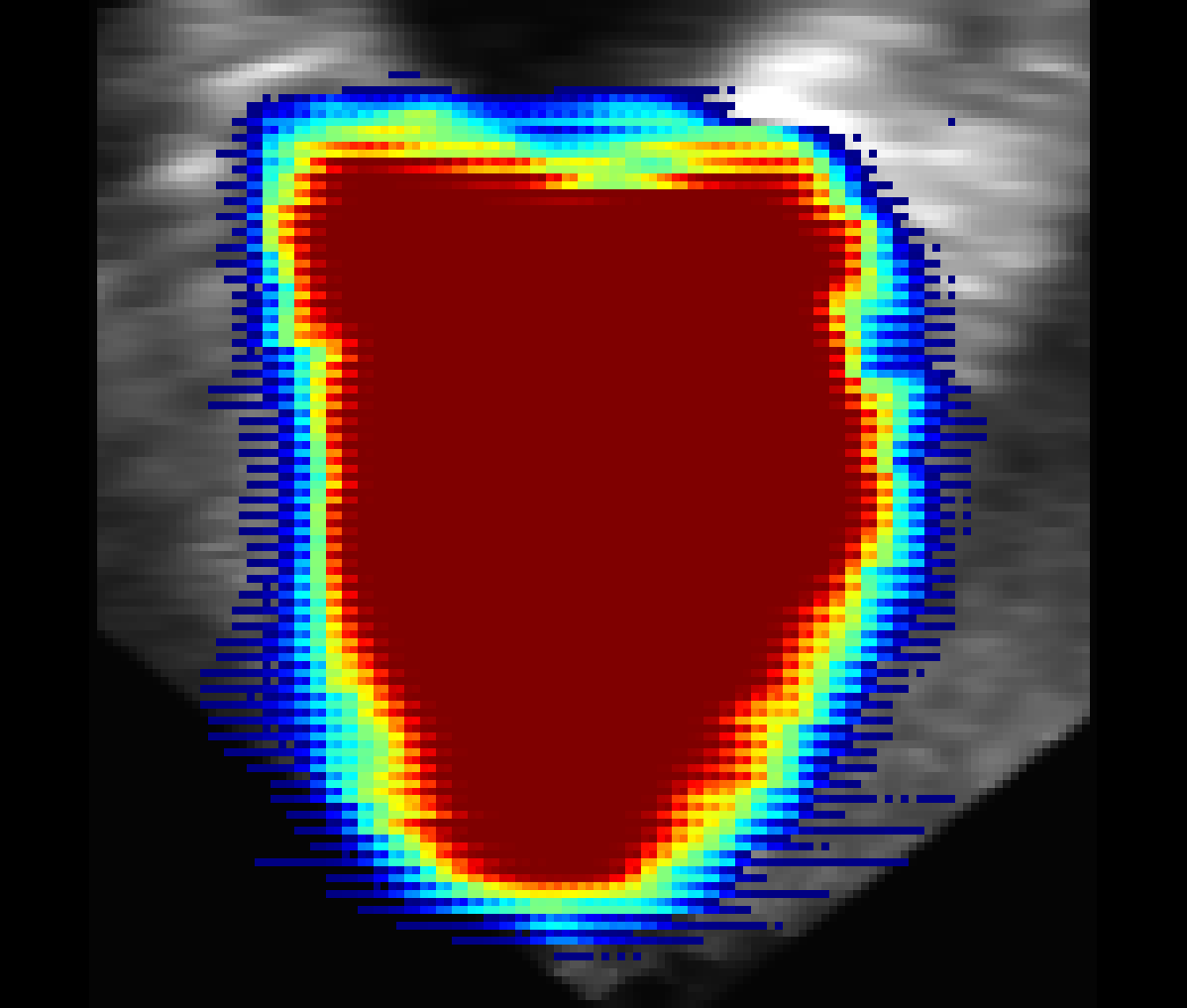}}
	\end{tabular}
	\caption{Our results. We show (a) the image, (b) overlaid (full) ground truth (used for evaluation only), (c) initial random walker prediction, and (d) our final segmentation result produced by the weakly supervised FCN. We show qualitative results for top to bottom: spleen (CT), liver (CT), prostate (MRI), and left ventricle (US) segmentation.
	\label{fig:results}}
\end{figure*}
\textbf{Datasets:} We utilize the training datasets (as they include ground truth annotations) from public challenges, specifically, from the \textit{Medical Segmentation Decathlon}\footnote{\url{http://medicaldecathlon.com}} and the \textit{Challenge on Endocardial Three-dimensional Ultrasound Segmentation}\footnote{\url{https://www.creatis.insa-lyon.fr/Challenge/CETUS/}}. All numbers are reported on 1 mm isotropic images that were generated from the original images using linear interpolation for both CT and MRI images. For ultrasound images, we keep their original resolution as they are close to isotropic. We employ random splits for training and validation for all datasets, resulting 32/9 cases for spleen (CT),  104/27 cases for liver (CT),  26/6 cases for prostate (MRI), and 24/6 cases for left ventricle (LV) in ultrasound (US).

\begin{figure}[htbp]
\begin{center}
    \adjincludegraphics[width=1.0\textwidth]{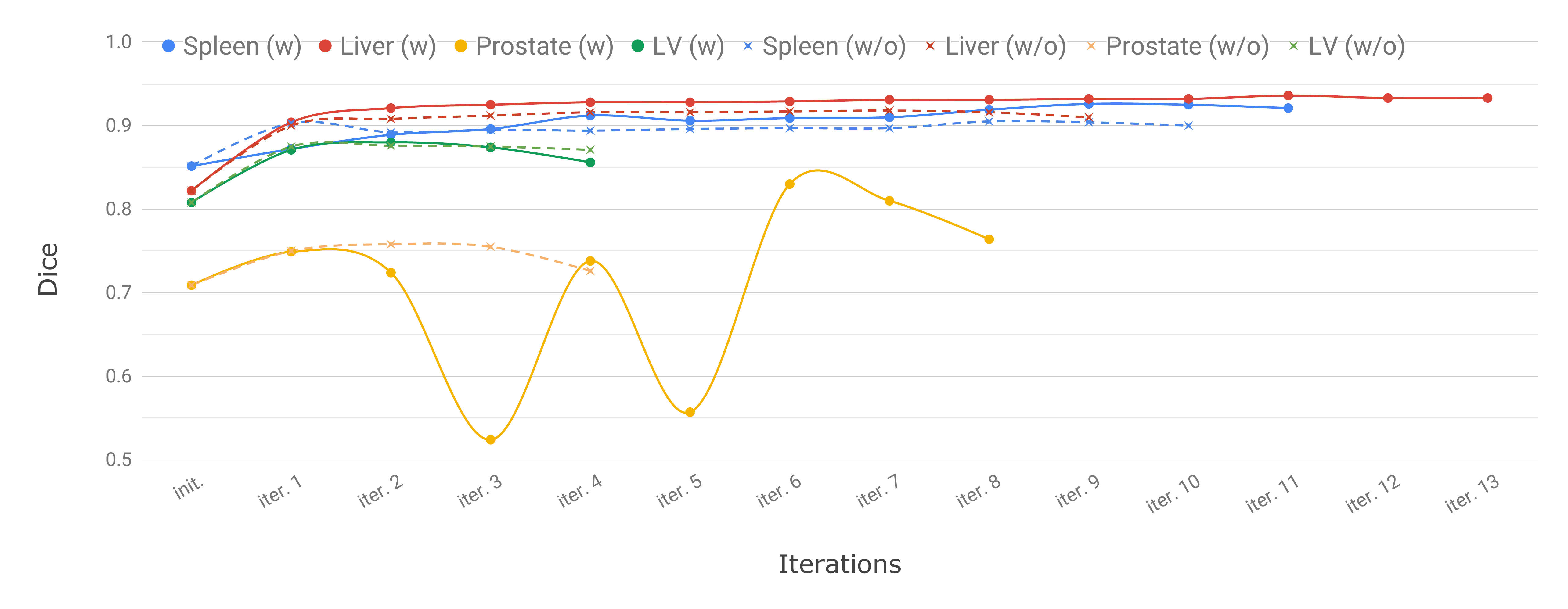}
\end{center}
\caption{Weakly supervised training from scribble based initialization. Each segmentation task is shown with (w) and without (w/o) random walker regularization after each round of FCN training.}
\label{fig:training}
\end{figure}

\textbf{Experiments:} In all cases, we iterate our algorithm until convergence on the validation data. We compare both training with and without employing random walker (RW) regularization after each round of 3D FCN training. Furthermore, we quantify the benefit of modelling the extreme points as an extra input channel to the network by running the framework with RW regularization but without the extreme points channel. The results are summarized in Table \ref{table:results} for all segmentation tasks. It can be observed that the biggest improvements happen in the first round FCN learning after initial random walker segmentation. While random walker regularization does not always improve the average Dice score, it does help to introduce enough ``novelty'' into our learning framework in order to drive the overall Dice score up in later iterations as shown in Fig. \ref{fig:training}. Visual examples of the improvement between from initial random walker to the final FCN prediction is shown in Fig. \ref{fig:results}.

\textbf{Implementation:} The training and evaluation of the deep neural networks used in the proposed framework were implemented based on the \textit{NVIDIA Clara Train SDK}\footnote{\url{https://devblogs.nvidia.com/annotate-adapt-model-medical-imaging-clara-train-sdk}} using NVIDIA Tesla V100 GPUs with 16 GB memory.
\begin{table}[htbp]
    \caption{Summary of our weakly supervised segmentation results. This table compares the random walker initialization with weakly supervised training from extreme points with (w) and without (w/o) random walker (RW) regularization, and with RW regularization but without the extra extreme points channel as input to the network (w RW; no extr.). For reference, the performance on the same task under fully supervised training is shown. \label{table:results}}
      \centering
      \footnotesize
\begin{tabular}{|l|l|l|l|l|}    
\hline
\textbf{Dice}& \textbf{Spleen (CT)}& \textbf{Liver (CT)}& \textbf{Prostate (MRI)}& \textbf{LV (US)}\\
\hline\hline
Rnd. walk. init.            & 0.852   & 0.822   & 0.709   & 0.808 \\
\hline
Weak. sup. (w/o RW)         & 0.905   & 0.918   & 0.758   & 0.876 \\
\hline
Weak. sup. (w RW; no extr.) & 0.924 &	0.935 &	0.779 &	0.860 \\
\hline
Weak. sup. (w RW)           & \textbf{0.926}   & \textbf{0.936}   & \textbf{0.830}   & \textbf{0.880} \\
\hline
\textbf{fully supervised}   & \textit{0.963}   & \textit{0.958}   & \textit{0.923}   & \textit{0.903} \\
\hline
\end{tabular}
\end{table}

\section{Discussion \& Conclusions}
\label{sec:conclusions}
We presented a method for weakly supervised 3D segmentation from extreme points. Asking the user to select the organ of interest using simple point clicks on the organ's surface in each spatial dimension can reduce the amount of labeling cost drastically. At the same time, the point clicks can describe the region of interest and simplify the machine learning task in 3D. Furthermore, the extreme points can be utilized to generate an initial weak pseudo label based on the extreme points utilizing the random walker algorithm. We found our initial label to be relatively robust to three diverse medical image segmentation tasks involving three different image modalities (CT, MRI, and ultrasound). Occasionally, the random walker can lack robustness for organs showing very diverse interior textures, like some advanced cancer patients in the prostate dataset. Here, a boundary search algorithm could potentially provide a better initial segmentation. Still, our FCN training in is able to markedly improve upon the initial segmentation.
Previous work mainly utilized bounding box annotations for weakly supervised learning, e.g. \cite{rajchl2017deepcut}. However, we consider selecting extreme points on the organ's surface to be more natural then selecting corners of a bounding box outside the organ of interest and more efficient than adding scribbles inside and around the organ \cite{wang2018deepigeos,can2018learning}. This is consistent to findings in the computer vision literature \cite{papadopoulos2017extreme}.
In the future, the region of interest and extreme point selection could be replaced by an automatic proposal network in order to further reduce the manual burden of medical image annotation.

\small
\bibliographystyle{splncs}

\end{document}